\documentclass[paper]{ieice}
\usepackage[pdftex]{graphicx,xcolor}
\usepackage[fleqn]{amsmath}
\usepackage{newtxtext}
\usepackage{cite}
\usepackage{amssymb,amsfonts}
\usepackage{algorithmic}
\usepackage{graphicx}
\usepackage{textcomp}
\usepackage{xcolor}
\usepackage{multirow}
\usepackage{subcaption}
\usepackage{color}
\field{A}

\field{}
\SpecialSection{ Enriched Multimedia - Advanced Safety, Security and Convenience - }
\title{Access Control with Encrypted Feature Maps for Object Detection Models}
\authorlist{%
 \authorentry{Teru Nagamori}{n}{labelA}\MembershipNumber{}
 \authorentry{Hiroki Ito}{n}{labelA}\MembershipNumber{}
 \authorentry{MaungMaung AprilPyone}{n}{labelA}\MembershipNumber{}
 \authorentry{Hitoshi Kiya}{f}{labelA}\MembershipNumber{7904021}
}
\affiliate[labelA]{The author is with the Tokyo Metropolitan University}

\received{2022}{3}{15}
\revised{2022}{8}{14}

\begin{document}
\maketitle
\begin{summary}
In this paper, we propose an access control method with a secret key for object detection models for the first time so that unauthorized users without a secret key cannot benefit from the performance of trained models. The method enables us not only to provide a high detection performance to authorized users but to also degrade the performance for unauthorized users. The use of transformed images was proposed for the access control of image classification models, but these images cannot be used for object detection models due to performance degradation. Accordingly, in this paper, selected feature maps are encrypted with a secret key for training and testing models, instead of input images. In an experiment, the protected models allowed authorized users to obtain almost the same performance as that of non-protected models but also with robustness against unauthorized access without a key.
\end{summary}

\section{Introduction}
Deep neural networks (DNNs) and convolutional neural networks (CNNs) have been deployed in many applications such as biometric authentication, automated deriving, and medical image analysis\cite{lecun2015deep}. However, training successful DNNs and CNNs requires three ingredients: a huge amount of data, GPU accelerated computing resources, and efficient algorithms, and it is not a trivial task. In fact, collecting images and labeling them is also costly and will also consume a massive amount of resources. Moreover, algorithms used in training a model may be patented or have restricted licenses. Therefore, trained DNNs and CNNs have great business value. Considering the expenses necessary for the expertise, money, and
time taken to train a model, a model should be regarded as a kind of intellectual property (IP).

There are two aspects of IP protection for DNN models: ownership verification and access control \cite{kiya2022overview}. The former focuses on identifying the ownership of the models, and the latter addresses protecting the functionality of the models from unauthorized access. Ownership verification methods were inspired by digital watermarking\cite{687830} and embed watermarks into models so that the embedded watermarks can be used to verify the ownership of the models in question\cite{xue2021dnn,uchida2017embedding,darvish2019deepsigns,10.1145/3323873.3325042,fan2019rethinking,lemerrer:hal-02264449,8695386,maung2021piracy,zhang2018protecting,217591}.

Although the above watermarking methods can facilitate in identifying the ownership of models, in reality, a stolen model can be exploited in many different ways. For example, an attacker can use a model for their own benefit without arousing suspicion, or a stolen model can be used for model inversion attacks \cite{10.1145/2810103.2813677} and adversarial attacks \cite{42503,43405,10.1145/3052973.3053009}. Therefore, it is crucial to investigate mechanisms to protect models from authorized access and misuse. In
this paper, we focus on protecting a model from misuse when it has been stolen (i.e., access control).

A method for protecting models against unauthorized access was inspired by adversarial examples and proposed to utilize secret perturbation to control the access of models \cite{chen2018protect}. In addition, another study introduced a secret key to protect models \cite{maungmaung_kiya_2021, 9291813}, and it was shown to outperform the other methods. The secret key-based protection method uses a key-based transformation that was originally used by an adversarial defense in \cite{adv-def}, which was in turn inspired by perceptual image encryption methods \cite{8448772,madono2020,8804201,8931606,9287532,SIP-113,1421853,8537968}. This block-wise model protection method utilizes a secret key in such a way that a stolen model cannot be used to its full capacity without a correct secret key. The robustness of these methods to attacks done to exploit protected models has already been verified, but all previous model-protection methods focus on the access control of image classification models.

Therefore, for the first time, in this paper, we propose a model protection method for object detection models by applying a key-based transformation to feature maps. The method not only achieves a high accuracy (i.e., almost the same accuracy as in the non-protected case), but also increases the key space substantially. We make the following contributions in this paper.
\begin{itemize}
\item We point out that the block-wise image encryption that is used for access control of image classification models is not useful for that of object detection models.
\item We propose an access control method with a secret key for object detection models for the first time, which enables us not only to maintain a high accuracy but also to increase the key space.
\end{itemize}
In an experiment, the proposed model-protection method is confirmed to outperform conventional methods with encrypted images.

\section{Related Work}
There are two approaches to protecting trained models: ownership verification
and access control. These approaches are reviewed to clarify problems with existing methods.
\subsection{Ownership Verification}
Ownership verification focuses on identifying the ownership of trained models. Researchers have developed model watermarking methods inspired by digital watermarking. In these methods, a watermark is embedded into a DNN model during the training phase, and the embedded watermark is used to claim ownership when the model is used without authorization.\par
There are two scenarios for the model watermarking methods: the white box scenario and the black box scenario.
The first model watermarking method was proposed in \cite{uchida2017embedding}, in which a watermark is embedded into one or more convolution layers of a model by using an embedding loss function. Similar methods were presented in \cite{darvish2019deepsigns,10.1145/3323873.3325042,fan2019rethinking}.
When extracting a watermark, these methods need access to model weights, so a white-box scenario is assumed.
However, trained models, including pirated ones, are often provided as an
online service. In such a situation, an inspector needs an ownership verification
method that uses only inputs and predictions, so a black-box scenario has to be considered. Several watermarking methods have been investigated to consider the black-box scenario such as in \cite{fan2019rethinking,lemerrer:hal-02264449,8695386,maung2021piracy,zhang2018protecting,217591}. In \cite{zhang2018protecting,217591}, a model owner installs a backdoor to output a predefined label for a particular input, like a kind  of adversarial example. Other methods extract a watermark pattern from the predictions of a protected model by using specific training samples \cite{fan2019rethinking,lemerrer:hal-02264449,8695386,maung2021piracy}.\par
However, model watermarking methods aim for only ownership verification. Therefore, stolen models can be directly used by unauthorized users, so we focus on access control to protect trained models from unauthorized access even if the models are stolen.

\begin{figure}[t]
    \centering
    \includegraphics[bb=0 0 721 653,scale=0.34]{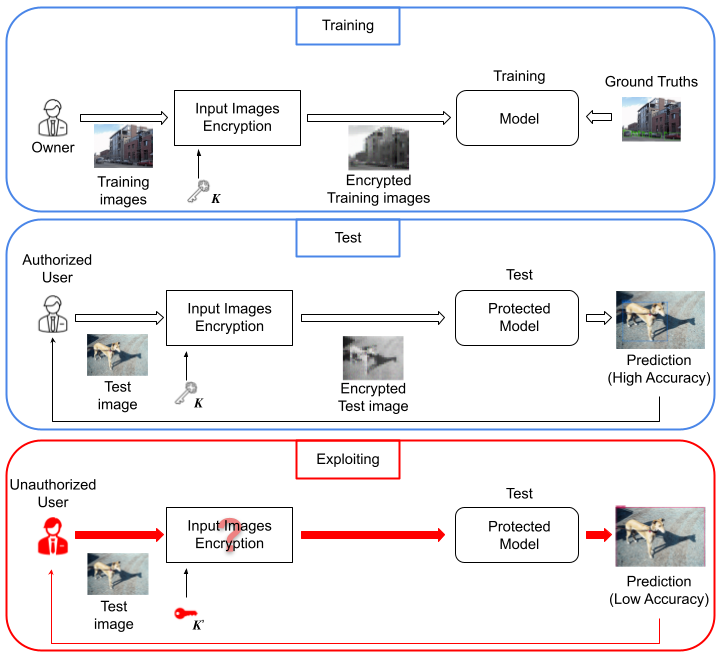}
    \caption{Access control framework using encrypted input images}
    \label{image_access_control}
\end{figure}

\begin{figure}[t]
    \centering
    \includegraphics[bb=0 0 721 653,scale=0.34]{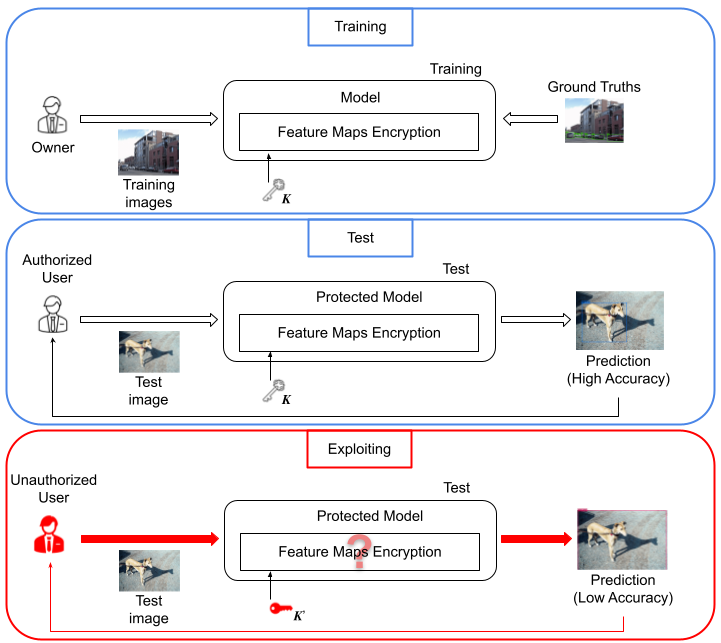}
    \caption{Access control framework using encrypted feature maps}
    \label{featuremap_access_control}
\end{figure}

\subsection{Access Control} 
Another approach to protecting trained models is access control that aims to protect the functionality of DNN models from unauthorized access. Therefore, protected models are required not only to provide a high accuracy to authorized users but also low performance to unauthorized users. In addition, since an unauthorized user may steal trained models to illegally use them, protected models have to be robust against various attacks.\par
The first access control method, which was inspired by adversarial examples \cite{42503,43405,10.1145/3052973.3053009}, was proposed for image classification models in \cite{chen2018protect}. In this method, authorized users add a secret perturbation generated by an anti-piracy transform module to input images, and the processed input images are fed to a protected model. Therefore, this method needs additional resources to train the module. In addition, the method focuses on protecting image classification models.\par
The second method is to extend the passport-based ownership verification method \cite{fan2019rethinking} as an access control method. However, the passport in \cite{fan2019rethinking} is a set of extracted features of a secret image/images or equivalent random patterns from a pretrained model. In addition, a network has to be modified with additional passport layers to use passports. Therefore, there are significant overhead costs in both the training  and inference phases. Moreover, the effectiveness of the passport-based method has never been confirmed under the use of object detection models.\par
The third is a block-wise image transformation method with a secret key \cite{maungmaung_kiya_2021, 9291813}, which is inspired by learnable image encryption \cite{adv-def,8448772,madono2020,8804201,8931606,9287532}. In this method, input images are encrypted with a key, for which three types of encryption methods: negative/positive   transformation (NP), pixel shuffling (SHF), and format-preserving Feistel-based encryption (FFX), were proposed.
Fig. \ref{image_access_control} shows the framework of the block-wise method. In the framework, an owner transforms all training images with secret key $K$, and a model is trained to protect the  model  by using  the  transformed  images  and  corresponding ground truths. An authorized user with key $K$ transforms a test image with K and feeds it to the protected model to get a prediction result with high accuracy. In contrast, an unauthorized user without key $K$ cannot obtain a prediction result with high accuracy, even if the unauthorized user knows the framework and the encryption algorithm. In addition, the method with key $K$ does not need any network modification or incur significant overhead costs. However, the use of the block-wise transformation is limited to image classification tasks, because the block-wise transformation cannot provide a pixel-level resolution that is required to protect object detection models. \par 
Accordingly, in this paper, we propose a novel access control method for object detection tasks for the first time. The proposed method does not need any network modification or incur significant overhead costs as well.

\section{Proposed Access Control with Encrypted Feature Maps}
Access control of object detection models with encrypted feature maps is
proposed here.

\subsection{Overview}
\begin{figure}[t]
    \centering
    \includegraphics[bb=0 0 719 219,scale=0.3]{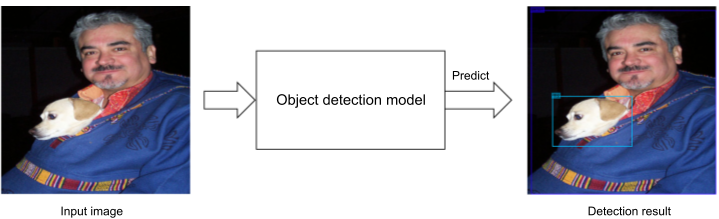}
    \caption{Overview of object detection}
    \label{detection}
\end{figure}

Protected models for access control should satisfy the following requirements.
The protected models should provide prediction results with a high accuracy to
authorized users but not provide such high-accuracy results to unauthorized
users. To satisfy these requirements, encrypted feature maps are used as shown
in Fig. \ref{featuremap_access_control}. 
In the framework with encrypted feature maps, an image owner trains a model
by using plain training images and corresponding ground truths, where selected
feature maps in the network are encrypted by using a secret key $K$ at each
training iteration in accordance with the proposed method. For testing, an
authorized user with key $K$ feeds a test image to the trained model to obtain a prediction result with high accuracy. In contrast, when an unauthorized user
without key $K$ inputs a test image to the trained model without any key or with an estimated key $K'$, the unauthorized user cannot benefit from the performance of the trained model.
The goal of object detection is to understand the regions and classes of objects
present in an image. Fig. \ref{detection} shows an example of object detection. Object detection detects the location of objects in the input image as rectangles, and classifies the class of each object. we used Single Shot MultiBox Detector 300 (SSD300)\cite{liu2016ssd} as an object detection model in this paper. \par

\begin{figure*}[htb]
    \centering
     \includegraphics[bb=0 0 1506 392,scale=0.33]{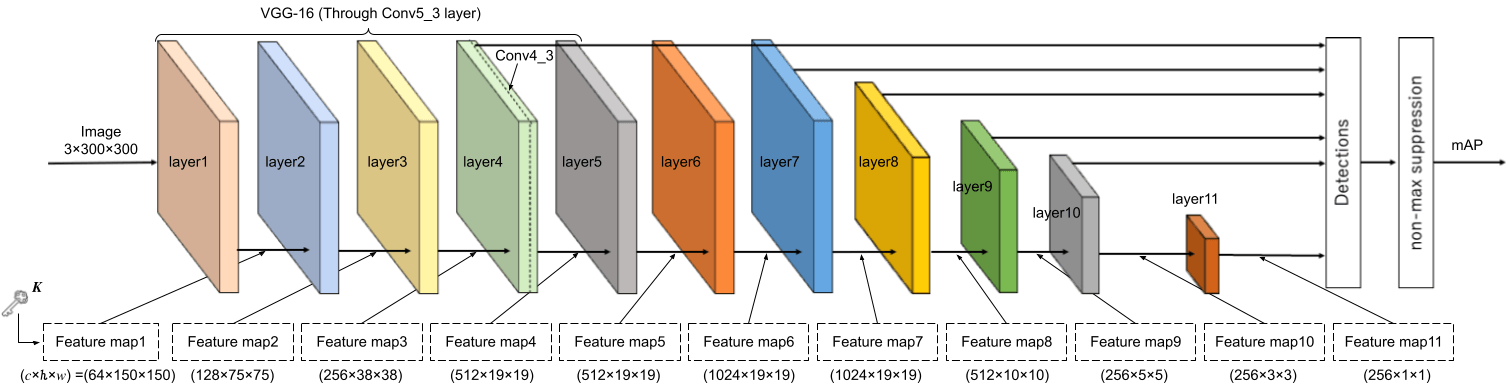}
    \caption{Architecture of object detection model (SSD300)}
    \label{proposed_model}
\end{figure*}

\begin{figure*}[tb]
    \centering
    \includegraphics[bb=0 0 989 273,scale=0.45]{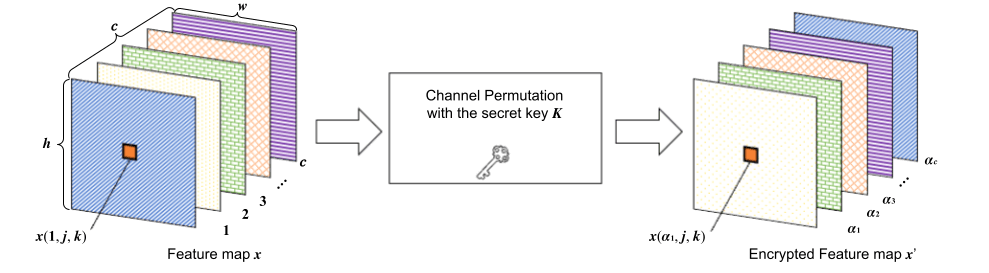}
    \caption{Feature map encryption (CP)}
    \label{transform}
\end{figure*}

\subsection{Feature Map}
In the proposed method, a feature map in a network is selected, and the
selected feature map is encrypted with a secret key. In this paper, SSD300\cite{liu2016ssd} based on VGG-16 \cite{simonyan2014very} is used as an object detection model, where SSD300 has 11 feature maps as illustrated in Fig. \ref{proposed_model}. In the proposed method, one feature map is selected from the feature maps, and it is encrypted with a key. \par 
As shown in Fig. \ref{proposed_model}, a feature map has a shape of ($c, h, w$), where $c$, $h$ and $w$ are the channel, height and width of the feature map, respectively. A color input image has $c$ = 3, but $c$ of a feature map is much larger in general. That is why we use a feature map.

\subsection{Feature Map Encryption}
In the conventional methods for image classification tasks, input images are encrypted with a key in accordance with a block-wise encryption method\cite{maungmaung_kiya_2021}. In contrast, in the proposed method, a feature map is encrypted in accordance with a pixel-wise encryption method so that results with a pixel-level resolution are obtained, because block-wise encryption cannot obtain such a resolution that is required to protect object detection models.\par
A selected feature map $x$ is transformed with a key $K$ at each iteration for training a model. To transform a feature map, pixel shuffling (SHF) used in image classification tasks is extended to be applied to object detection models in terms of two points: the use of feature maps and a block size of $M$ = 1. The extended encryption is called channel permutation (CP). Accordingly, CP is a pixel-wise transformation, where a feature map is permuted only along the channel dimension (see Fig. \ref{transform}). The following is the procedure of CP.
 
\begin{itemize}
    \item[1)]Select a feature map $x$ to be encrypted.
    \item[2)]Split $x$ into blocks with a size of $M$ as
        \begin{equation}
            B = \left\{B_{(1,1)},\ldots,B_{(i,j)},\ldots,B_{(h/M,w/M)}\right\} , 
        \end{equation}
        where $h$ and $w$ are assumed to be dividable by $M$ for the sake of simplicity, and $M$ = 1 is chosen for CP.
    \item[3)]Flatten each block $B_{(i,j)}$ into a vector as
        \begin{equation}
            b_{(i,j)} = [b_{(i,j)}(1),\ldots,b_{(i,j)}(L)] , 
        \end{equation}
    where $L$ is $c \times M \times M$.
    
    \item[4)]Generate a secret key as
        \begin{equation}
            \alpha = [\alpha_1,.,\alpha_i,...,\alpha_{i'},...,\alpha_L], 
        \end{equation}
         where $\alpha_i \in \left\{1,...,L\right\}$ and { $\alpha_i \ne \alpha_{i'}$ } if $i\ne i'$.\par
   
    \item[5)]Permutate each vector $b_{(i,j)}$ with $\alpha$ as
        \begin{equation}
                b'_{(i,j)}(i) = b_{(i,j)}(\alpha_i),
            \end{equation}
            and get a permutated vector such as
            \begin{equation}
                b'_{(i,j)} = [b'_{(i,j)}(1),\ldots,b'_{(i,j)}(L)].
            \end{equation}
            
    \item[6)]Concatenate the permutated vectors to form a encrypted feature map $x'$ with a dimension of ($c, h, w$).
    
\end{itemize}

\begin{figure*}[t]
    \centering
    \includegraphics[bb=0 0 989 273,scale=0.45]{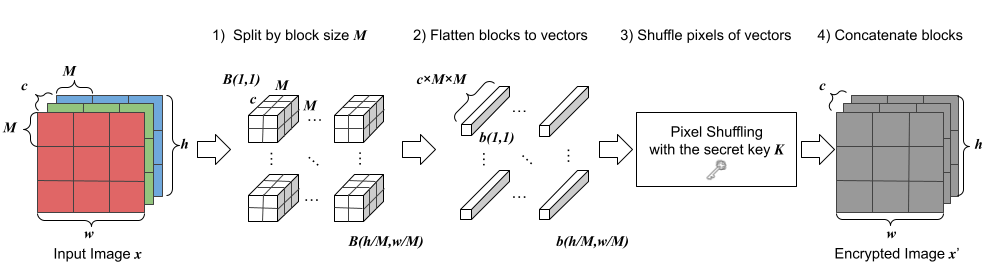}
    \caption{Input image encryption\cite{maungmaung_kiya_2021}}
    \label{input_image_transform}
\end{figure*}

\begin{figure*}[t]
\centering
        \begin{minipage}{4truecm}
             \centering
              \includegraphics[bb=0 0 542 542,scale=0.2]{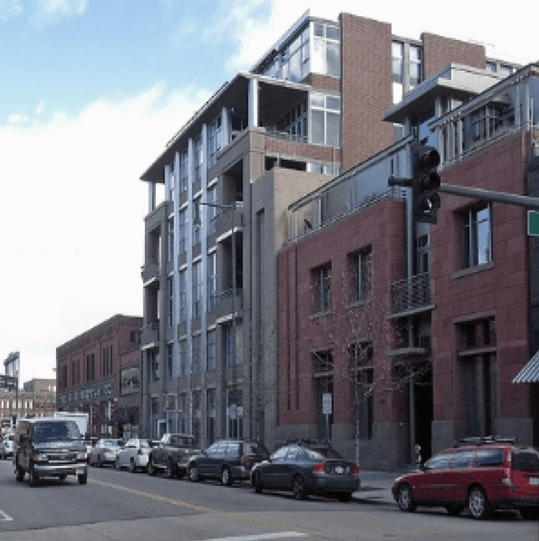}
              \subcaption{Original}
            \end{minipage}
        \begin{minipage}{4truecm}
             \centering
              \includegraphics[bb=0 0 541 542,scale=0.2]{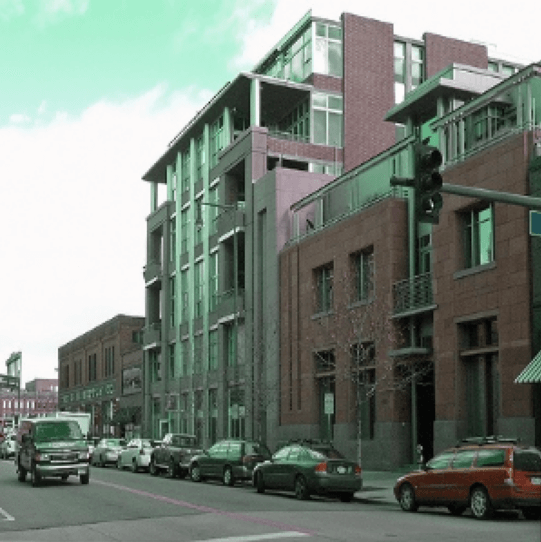}
              \subcaption{$M$=1}\label{blocksize1}
            \end{minipage}
        \begin{minipage}{4truecm}
             \centering
              \includegraphics[bb=0 0 543 543,scale=0.2]{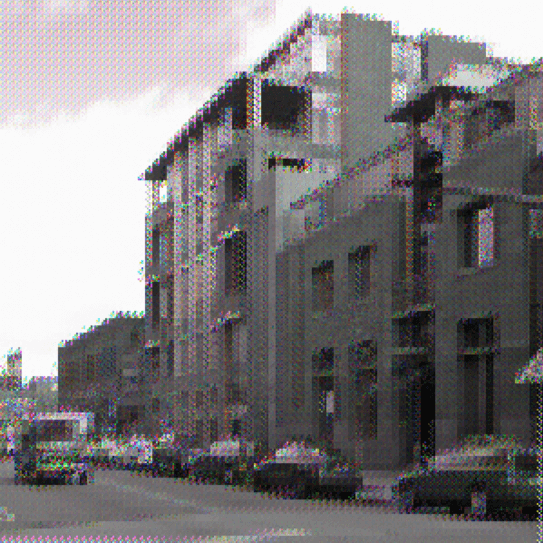}
              \subcaption{$M$=4}
            \end{minipage}
        \\
        \begin{minipage}{4truecm}
             \centering
              \includegraphics[bb=0 0 542 541,scale=0.2]{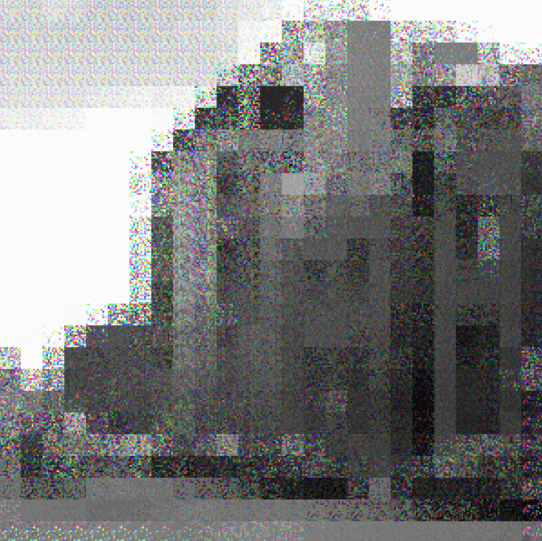}
              \subcaption{$M$=12}
            \end{minipage}
        \begin{minipage}{4truecm}
             \centering
              \includegraphics[bb=0 0 542 542,scale=0.2]{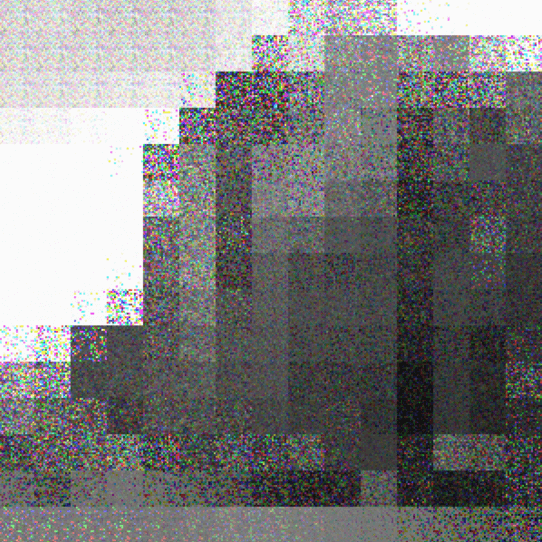}
              \subcaption{$M$=20}
            \end{minipage}
        \begin{minipage}{4truecm}
            \centering
              \includegraphics[bb=0 0 537 538,scale=0.2]{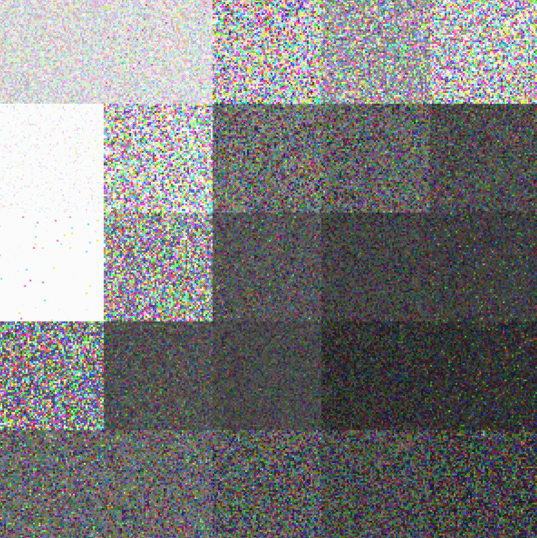}
              \subcaption{$M$=60}
            \end{minipage}\\
    \caption{Examples of encrypted image with pixel shuffling ($M$ = block size)}
    \label{encrypted_iamges}
\end{figure*}

The above procedure corresponds to that of a block-wise image encryption used in image classification tasks \cite{maungmaung_kiya_2021}, if $x$ is an input image, and $M$ is larger than $M$ = 1, as illustrated in Fig. \ref{input_image_transform}. Fig. \ref{encrypted_iamges} also shows an example of images transformed by using the bock-wise encryption.
 
\begin{table}[t]
 \caption{Key space of encryption methods}
 \label{table:key space}
 \centering
  \begin{tabular}{|c|cc|}
   \hline
   Encryption Method  & Key space & Remark\\
   \hline
   SHF\cite{maungmaung_kiya_2021} & $(c \times M \times M)!$ & $c =$ 3  \\
   CP & $c!$ & $c >$ 3 \\
   \hline
  \end{tabular}
\end{table}

\subsection{Difference between SHF and CP}
The differences between SHF for image classification models and CP for object detection ones are summarized as below.
\begin{itemize}
\item [(a)] CP is a pixel-wise transformation.
\item [(b)] CP is applied to a feature map.
\item [(c)] The number of feature map channels is larger than that of input image channels.
\end{itemize}
Difference (a) allows us to obtain results with a pixel-level resolution, but pixel-wise transformations are not robust against various attacks if the transformation is applied to input images as discussed in \cite{maungmaung_kiya_2021} because the number of input image channels $c$ is small (i.e., RGB images have $c$ = 3). \par
In Table \ref{table:key space}, the key space of SHF is compared with that of CP. For SHF, the key is decreased if a small bock size is chosen. For example, if SHF with $M$ = 1 is applied to an input image, the key space is $3!$ = 6 . In contrast, when a feature map with $c$ = 256 is encrypted by using CP, the key space is $256! \approx 2^{1684}$. In general, the channel number of feature maps is much larger than that of input images, so CP has a large key space even when
$M$ = 1 is selected, compared with SHF. In addition, when using CP, the attacker has to know or estimate the location of the transformed feature map, which cannot be known from the model itself. In our scenario, the model owner is assumed to securely manages both the location of the transformed feature map and the key. Accordingly, the attacker cannot know the location of the transformed feature map from the network.

\subsection{Threat model}
A threat model includes a set of assumptions such as an attacker’s goals, knowledge, and capabilities. Users without secret key $K$ are assumed to be the adversary. Attackers may steal a model to achieve different goals for profit. In this paper, we consider the attacker’s goal is to be able to make use of a stolen model.\par
We assume that authorized users know key $K$, and the model owner securely manages both the location of transformed feature maps and key $K$. In addition, the model protection method is also assumed to be disclosed except for key $K$ and the location of transformed feature maps. 
To use stolen models, the adversary has to get all of the location of transformed feature maps, key $K$ and the algorithm with $K$ for testing a query image. Accordingly, if the adversary can get these three items, it is possible to use the models, but the adversary does not know the location of transformed feature maps and the algorithm for testing a query image in general even in the case of a malicious legitimate user or collusion attacks. However, the adversary can be a spoofed-authorized user if he/she gets key $K$ from a malicious legitimate user or in some way. As a result, in that case, the adversary can illegally use the protected model.

\section{Experimental Results}
To confirm the effectiveness of the proposed method, we evaluated it in terms of access control and robustness against unauthorized access. All the experiments were conducted using the PyTorch library in Python \cite{NEURIPS2019_bdbca288}.

\subsection{Setup}
As an object detection model, we used the SSD300 pre-trained by using the ImageNet Large-Scale Visual Recognition Challenge (ILSVRC) CLS-LOC dataset \cite{russakovsky2015imagenet}. To apply the model to an object detection task, the pre-trained model was then fine-tuned by using two datasets.

\subsubsection{PASCAL VOC}
The PASCAL visual object classes (VOC) challenge 2007 \cite{everingham2010pascal} and 2012 \cite{everingham2015pascal} trainval datasets were used to fine-tune the pre-trained model. The PASCAL VOC 2007 was also used for testing. These datasets have 20 categories, including people, dogs, and so on.
For data augmentation, the random sample crop, horizontal flip, and some photometric distortions described in \cite{liu2016ssd} were used for training models. In addition, due to the restrictions of SSD300 shown in Fig. \ref{proposed_model}, input images were resized to $300\times300$ pixels.\par
Models were trained by using a stochastic gradient descent (SGD) optimizer with an initial learning rate of $10^{-\:3}$, a momentum value of 0.9, a weight decay value of 0.0005, and a batch size of 32.
Models were trained for 60k iterations with a learning rate of $10^{-\:3}$. The models were then trained for 20k iterations with a learning rate of $10^{-\:4}$ and 40k iterations with a learning rate of $10^{-\:5}$.
The overall objective loss function was a weighted sum of the localization loss and the confidence loss. 
In this paper, the confidence loss was the cross-entropy loss over multiple classes confidences, and the localization loss was the Smooth L1 loss between the predicted position and the ground truth position.

\subsubsection{MS COCO}
Another dataset called MS COCO was also used to fine-tune the pre-trained model. The pre-trained model was fine-tuned by using the COCO2014 train dataset \cite{COCO}. 10,000 images, which were randomly selected from the COCO2014 validation dataset \cite{COCO}, were used as test images. In object detection tasks, the datasets have 80 categories, so they are more difficult than the PASCAL VOC dataset.\par 
In the experiment, we used the same data augmentation, model structure, and losses as the PASCAL VOC dataset except for the number of model iterations. Models were trained for 160k iterations at a learning rate of $10^{-\:3}$. The models were then trained 40k iterations at a learning rate of $10^{-\:4}$ and 40k iterations at a learning rate of $10^{-\:5}$.

\subsection{Metric for Evaluating Detection Performance}
Mean average precision (mAP) \cite{liu2016ssd} was used as a metric for evaluating detection performance, which is a common evaluation metric for object detection. In this paper, the mAP of the PASCAL VOC dataset was calculated in accordance with the calculation method used in the PASCAL VOC 2007 \cite{everingham2010pascal} as follows.\par
First, intersection over union (IoU) \cite{rezatofighi2019generalized} for each object is calculated by
\begin{equation}
    IoU = \frac{A_{truth}\cap A_{pred}}{A_{truth}\cup A_{pred}},
\end{equation}
where $A_{truth}$ represents the rectangular area of a ground truth, and $A_{pred}$ is a
predicted rectangular area. Precision and recall are then defined for each class label
as
\begin{align}
    Precision =& \frac{TP}{TP+FP}\\
    Recall =&  \frac{TP}{TP+FN} ,
\end{align}
 where $TP$ (true positive) is the number of predicted objects for which there exists a ground truth object that satisfy two conditions: the same class label as
that of the ground truth object, and $IoU \geq 0.5$. $FP$ (false positive) is the number of predicted objects that do not satisfy the conditions, and
$FN$ (false negative) is the number of ground truths that are not detected. \par
Then, average precision (AP) for each class is calculated by using precision and recall in Eqs. (7) and (8).
Finally, a mean average precision (mAP) value is calculated as
\begin{equation}
    mAP = \frac{1}{N}\sum_{class=1}^{N}AP_{class} ,
\end{equation}
where $N$ is the number of classes and $AP_{class}$ is the AP value of each class. 
The mAP ranges from 0 to 1, where when a mAP value is closer to 1, it indicates a higher accuracy.

In addition, we used the Python version of COCO API published by MS COCO to evaluate the MS COCO dataset. Among the results obtained from COCO API, mean average precision (mAP) under three different values of IoU (0.5, 0.75, 0.5:0.95) was used in the experiment, where 0.5:0.95 represents taking the average of the AP values when a value of IoU is changed from 0.5 to 0.95 in 0.05 increments.

\begin{figure*}[t]

\scalebox{0.8}[0.8]{
    
    \begin{tabular*}{50mm}{@{\extracolsep{\fill}}c|c|ccc}
        Ground Truth&Baseline&Correct ($K$)&No-enc&Incorrect ($K'$) \\
        \begin{minipage}{4truecm}
             \centering
              \includegraphics[bb=0 0 393 276,scale=0.25]{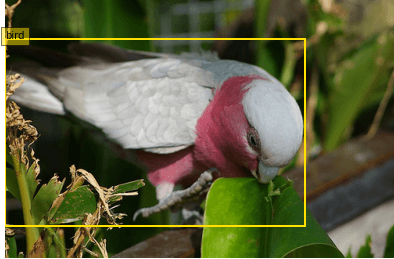}
            \end{minipage}
        &
        \begin{minipage}{4truecm}
             \centering
              \includegraphics[bb=0 0 393 276,scale=0.25]{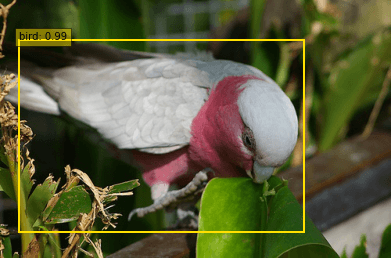}
            \end{minipage}
        &
        \begin{minipage}{4truecm}
             \centering
              \includegraphics[bb=0 0 393 276,scale=0.25]{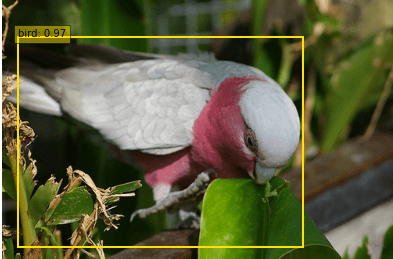}
            \end{minipage}
        &
        \begin{minipage}{4truecm}
             \centering
              \includegraphics[bb=0 0 393 276,scale=0.25]{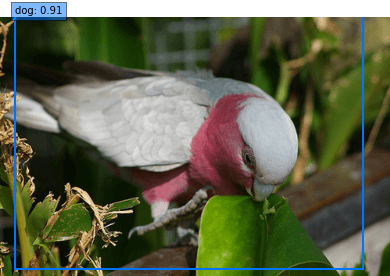}
            \end{minipage}
        &
        \begin{minipage}{4truecm}
            \centering
              \includegraphics[bb=0 0 393 276,scale=0.25]{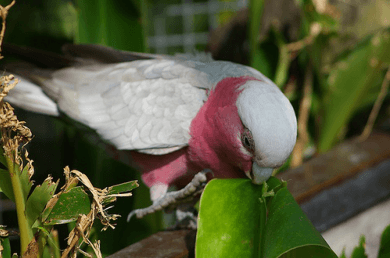}
            \end{minipage}\\
        \begin{minipage}{4truecm}
             \centering
              \includegraphics[bb=0 0 395 293,scale=0.25]{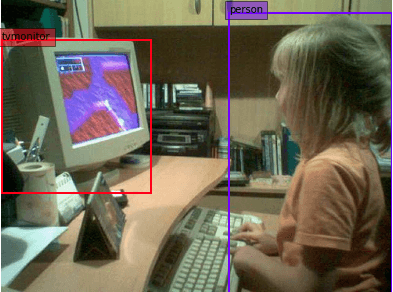}
            \end{minipage}
        &
        \begin{minipage}{4truecm}
             \centering
              \includegraphics[bb=0 0 395 293,scale=0.25]{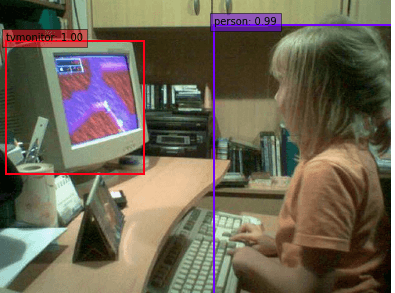}
            \end{minipage}
        &
        \begin{minipage}{4truecm}
             \centering
              \includegraphics[bb=0 0 395 293,scale=0.25]{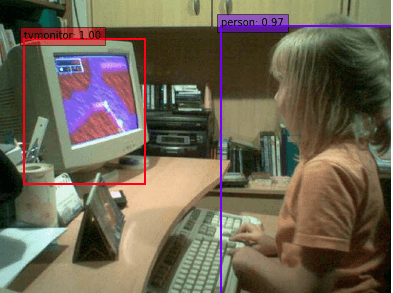}
            \end{minipage}
        &
        \begin{minipage}{4truecm}
             \centering
              \includegraphics[bb=0 0 395 293,scale=0.25]{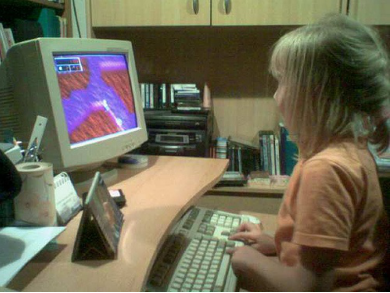}
            \end{minipage}
        &
        \begin{minipage}{4truecm}
             \centering
              \includegraphics[bb=0 0 395 293,scale=0.25]{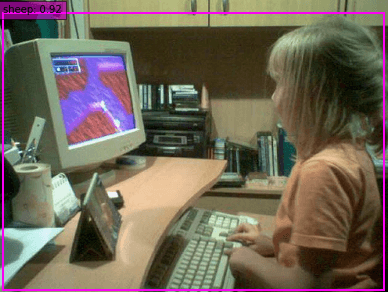}
            \end{minipage}\\
        \begin{minipage}{4truecm}
             \centering
              \includegraphics[bb=0 0 400 309,scale=0.25]{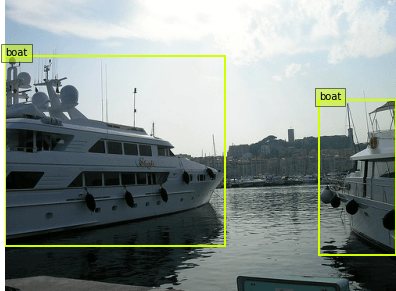}
            \end{minipage}
        &
        \begin{minipage}{4truecm}
             \centering
              \includegraphics[bb=0 0 400 309,scale=0.25]{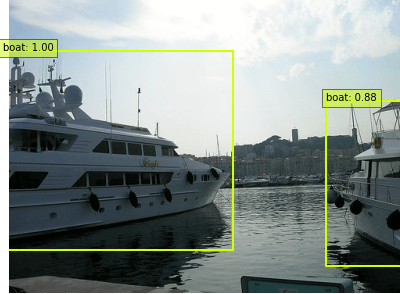}
            \end{minipage}
        &
        \begin{minipage}{4truecm}
              \centering  
              \includegraphics[bb=0 0 400 309,scale=0.25]{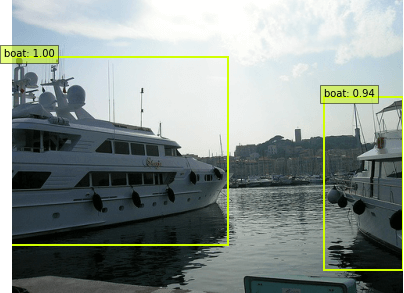}
            \end{minipage}
        &
        \begin{minipage}{4truecm}
             \centering
              \includegraphics[bb=0 0 400 309,scale=0.25]{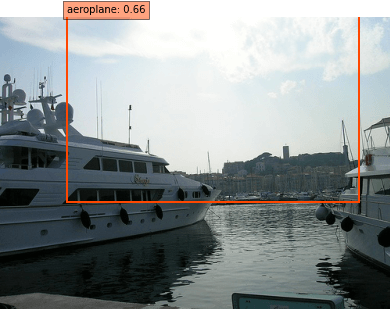}
            \end{minipage}
        &
        \begin{minipage}{4truecm}
            \centering
              \includegraphics[bb=0 0 400 309,scale=0.25]{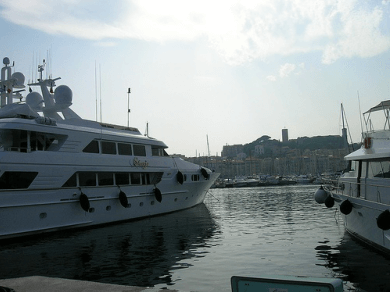}
            \end{minipage}\\
        
    \end{tabular*}
    }
    \caption{Example of detected objects (PASCAL VOC dataset, Model-4)}
    \label{detection_result}
\end{figure*}

\subsection{Detection Result under Encrypted Input Images}
First, input images were encrypted in accordance with SHF under various block sizes (i.e., $M$ $\in \left\{1, 4, 12, 20, 60\right\}$) for comparison with the proposed method (CP) on the PASCAL VOC dataset. SHF was already demonstrated to achieve a high access control performance in image classification tasks in \cite{maungmaung_kiya_2021}, but it has never been applied to object detection ones.

Table \ref{table:SHF} shows experiment results, where "Correct ($K$)" means the use of test images encrypted with correct key $K$, and “No-enc” indicates the use of plain test images. “Baseline” also indicates the use of plain test images and models trained with plain images. “Incorrect ($K'$)" means the use of randomly generated key $K'$, where each value of Incorrect ($K'$) was calculated by averaging the values of 100 tests. From the results, even when correct key $K$ was used, the accuracy decreased significantly as block size $M$ increased. In contrast, when the block size was small, the protected model achieved a high accuracy close to the baseline. However, the accuracy without the encryption (i.e., No-enc) or with “ Incorrect” was almost the same as that of “Correct,” so the access control was weak under the use of a small block size due to a small key space.

\begin{table}[tb]
 \caption{Detection accuracy (mAP) of models with \\ encrypted input images (PASCAL VOC dataset)}
 \label{table:SHF}
 \centering
  \scalebox{0.83}{
    \begin{tabular}{|c|c|ccc|}
       \hline
       method & block size  & Correct ($K$) & No-enc & Incorrect ($K'$) \\
       \hline
       \multirow{5}{*}{pixel shuffling (SHF)} & 1 & 0.7710 & 0.7598 & 0.7603 \\
       &4& 0.7154 & 0.5745 & 0.3883 \\
       &12&  0.4891 & 0.1976 & 0.0910 \\
       &20& 0.0083 & 0.0086 & 0.0065 \\
       &60& 0.1284 & 0.0480 & 0.0416 \\
       \hline
       \multicolumn{2}{|c|}{Baseline} & \multicolumn{3}{c|}{0.7690}\\
       \hline
    \end{tabular}
  }
\end{table}
\subsection{Detection Result under Encrypted Feature Maps}
A selected feature map was transformed with key $K$ in accordance with the procedure in sec. 3.3. Table \ref{table:result} shows experimental results on the PASCAL VOC dataset, where Model-1 means that feature map 1 was selected for encryption. Figure \ref{detection_result} also shows an example of detected objects, in which feature map 4 was encrypted.
From Table \ref{table:result} and Fig. \ref{detection_result}, the proposed method (CP) was confirmed to achieve almost the same accuracy as that of the baseline under the use of the correct key when feature map 2, 4, 5, 6, or 7 was selected. In contrast, CP provided a low accuracy to unauthorized users both without the key (No-enc) and with an incorrect key.\par
Table \ref{table:COCOresult} also shows experimental results of using the proposed method on the MS COCO dataset where we focused on six models from  Model-1 to Model-6, which achieved a high performance on the PASCAL VOC dataset.
From the table, the proposed method was confirmed to be able to have almost same performance as that of Baseline as well as the use of the PASCAL VOC dataset, when correct key $K$ was used. In contrast, when without correct key $K$, the accuracy was heavily degraded as well.\par
Accordingly, CP with an encrypted feature map was effective in the access control of object detection models.

\begin{table}[bt]
 \caption{Detection accuracy (mAP) of proposed models (PASCAL VOC dataset)}
 \label{table:result}
 \centering
  \begin{tabular}{|c|ccc|}
   \hline
   Selected feature map  & Correct ($K$) & No-enc & Incorrect ($K'$) \\
   \hline
   Model-1& 0.7244 & 0.1363 & 0.0421 \\
   Model-2& 0.7611 & 0.0091 & 0.0180 \\
   Model-3& 0.7475 & 0.0091 & 0.0078 \\
   Model-4& 0.7611 & 0.0023 & 0.0043 \\
   Model-5& 0.7587 & 0.1672 & 0.1624\\
   Model-6& 0.7617 & 0.1732 & 0.1672\\
   Model-7& 0.7695 & 0.1768 & 0.1750 \\
   Model-8& 0.7677 & 0.3529 & 0.3415\\
   Model-9& 0.7705 & 0.5767 & 0.5678 \\
   Model-10& 0.7705 & 0.7177 & 0.7027 \\
   Model-11& 0.7512 & 0.7314 & 0.7252 \\
   \hline
   Baseline& \multicolumn{3}{c|}{0.7690}\\
   \hline
  \end{tabular}
\end{table}

\begin{table*}[bt]
 \caption{Detection accuracy (mAP) of proposed models (COCO dataset)}
 \label{table:COCOresult}
 \centering
 \scalebox{0.8}[0.8]{
  \begin{tabular}{|c|c|ccc|ccc|ccc|}
   \hline
   \multicolumn{1}{|c}{} &  & \multicolumn{3}{c|}{0.5:0.95} & \multicolumn{3}{c|}{0.5} & \multicolumn{3}{c|}{0.75} \\
   \cline{3-11}
    \multicolumn{1}{|c}{} &    & Correct ($K$) & No-enc & Incorrect ($K'$) & Correct ($K$) & No-enc & Incorrect ($K'$) & Correct ($K$) & No-enc & Incorrect ($K'$)\\
    \hline
    \multirow{6}{*}{Selected feature map}& 
      Model-1 & 0.149 & 0.020 & 0.002 & 0.272 & 0.041 & 0.004 & 0.149 & 0.018 & 0.001\\ 
    & Model-2 & 0.139 & 0.004 & 0.000 & 0.253 & 0.009 & 0.001 & 0.138 & 0.003 & 0.000\\ 
    & Model-3 & 0.135 & 0.000 & 0.000 & 0.249 & 0.001 & 0.000 & 0.135 & 0.000 & 0.000\\
    & Model-4 & 0.149 & 0.045 & 0.045 & 0.271 & 0.102 & 0.101 & 0.148 & 0.036 & 0.036\\ 
    & Model-5 & 0.152 & 0.046 & 0.046 & 0.277 & 0.105 & 0.105 & 0.152 & 0.036 & 0.036\\ 
    & Model-6 & 0.152 & 0.047 & 0.047 & 0.276 & 0.107 & 0.107 & 0.152 & 0.038 & 0.038\\ 
    \hline
    \multicolumn{2}{|c|}{Baseline} & \multicolumn{3}{c|}{0.150} & \multicolumn{3}{c|}{0.272} & \multicolumn{3}{c|}{0.151}\\
   \hline
   
  \end{tabular}
  }
\end{table*}

\subsection{Robustness against Random Key Attack}
\begin{figure}[t]
    \centering
    \includegraphics[bb=0 0 432 288,scale=0.55]{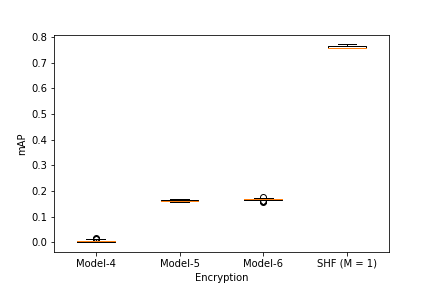}
    \caption{mAP values of protected models with 100 incorrect keys (PASCAL VOC dataset). Boxes span from first to third quartile, referred to as $Q_{1}$ and $Q_{3}$, and whiskers show maximum and minimum values in range of [$Q_{1} - 1.5(Q_{3} - Q_{1}), Q_{3} + 1.5(Q_{3} - Q_{1})$]. Band inside box indicates median. Outliers are indicated as dots. }
    \label{boxplot}
\end{figure}

CP was evaluated in terms of robustness against the random key attack that is to use test images encrypted with a randomly generated key. An evaluation was carried out on robustness with 100 incorrect keys.
Fig. \ref{boxplot}  shows the detection performance of the protected models under the use of the incorrect keys on the PASACAL VOC dataset. From the results of using CP, the mAP values were significantly low for all selected feature maps: Model-4, Model-5 and Model-6, so the models were robust enough against this attack. Also, the mAP values of using SHF were high. Therefore, the proposed method (CP) outperformed the conventional method (SHF) in terms of robustness against the random key attack.\par 
From Table \ref{table:result}, the performance of models trained with an encrypted feature map depended on the selection of feature maps. We think that the difference in accuracy among the selected feature maps was caused by the architecture of the SSD300. From Fig. \ref{proposed_model}, the SSD uses multiple layers for feature learning. Therefore, for Model-10, only layer 11 is affected by encryption, so layers 4, 7, 8, and 9 can use almost the same features as the baseline model. In other words, the use of a deeper feature map makes the influence of CP weaker.

\subsection{Impact of CP on training}
Model-5 with CP was compared with Baseline in terms of the time required for each iteration and the speed of learning convergence. The average time of Baseline was 0.1670 sec and that of Model-5 was 0.2169 sec for each iteration where the machine specs used for evaluating the execution time were listed in Table \ref{table:machine info}. The use of CP was confirmed to slightly increase the time required for each iteration.\par
In Table \ref{table:loss value}, the speed of learning convergence with CP was compared with that without CP (Baseline). From the table, the learning convergence of the method with CP was almost the same as that without CP.

\begin{table}[bt]
 \caption{Machine spec used for evaluating executing time}
 \label{table:machine info}
 \centering
  \begin{tabular}{|c|c|}
  \hline
   Processor & Intel(R) Core(TM) i7-4790K CPU @ 4.00GHz\\
   OS & Ubuntu 18.04.6 LTS\\
   GPU & Quadro RTX 6000\\
   Memory (CPU) &  32GB\\
   Memory (GPU) &  24GB\\
  \hline
  \end{tabular}
\end{table}

\begin{table}[bt]
 \caption{Learning convergence speed with and without CP (PASCAL VOC dataset)}
 \label{table:loss value}
 \centering
  \begin{tabular}{|c|cc|}
  \hline
   &\multicolumn{2}{c|}{Loss value} \\
   \cline{2-3}
   No. of iterations  & Baseline & Model-5 \\
   \hline
   20000 & 3.5232 & 3.2191 \\
   40000& 2.9450 & 3.1328  \\
   80000& 2.3215 & 2.2475 \\
  120000& 1.8859 & 2.0459 \\
  \hline
  \end{tabular}
\end{table}

\section{Conclusion}
In this paper, we proposed an access control method for object detection models for the first time. The method is carried out by encrypting a selected feature map with a secret key called channel permutation (CP), while input images are encrypted by using a block-wise encryption method in conventional methods. The use of CP allows us not only to obtain a pixel-level accuracy that is required for object detection but also to maintain a large key space even when a pixel-wise permutation is used. As a result, the proposed method can maintain both a high accuracy and robustness against attacks. In experiments, the conventional method with encrypted input images was not effective in the access control of object detection models, and the effectiveness of the proposed method was demonstrated in terms of detection accuracy. As future work, the proposed method with a encrypted feature map will be applied to other models to verify the effectiveness and limitations of CP.

\section*{Acknowledgement}
This study was partially supported by JSPS KAKENHI (Grant Number JP21H01327).

\bibliographystyle{ieicetr} 
\bibliography{ref} 

\profile[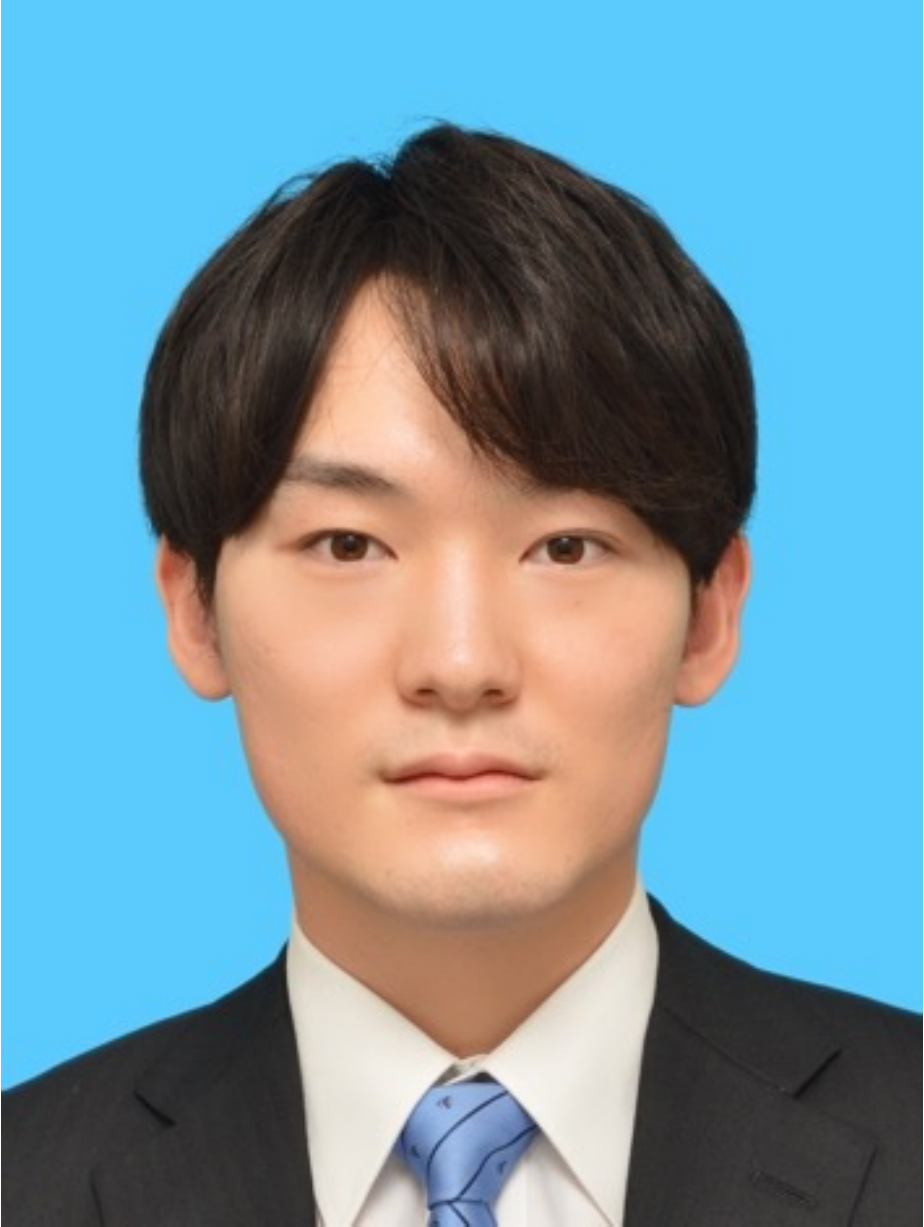]{Teru Nagamori}{received his B.C.S degree from the Tokyo Metropolitan University, Japan in 2022. Since 2022, he has been a Master course student at the Tokyo Metropolitan University. His research interests include deep neural networks and their protection.}

\profile[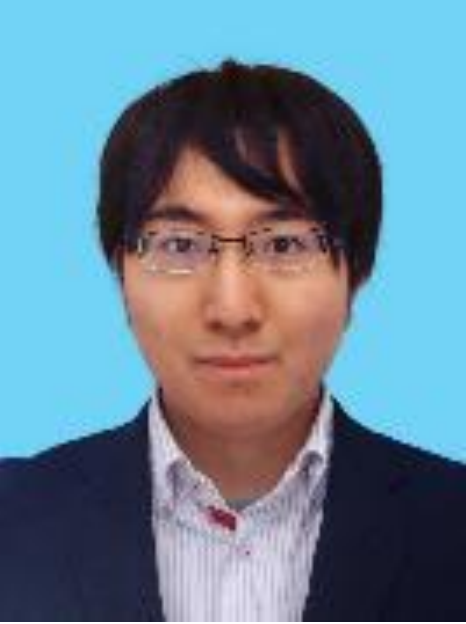]{Hiroki Ito}{received his B.Eng.degree from the Tokyo Metropolitan University, Japan, in 2020. From 2020, he has been a Master course student at the Tokyo Metropolitan University. His research interests include deep neural networks and their protection.}
\vspace{5mm}
\profile[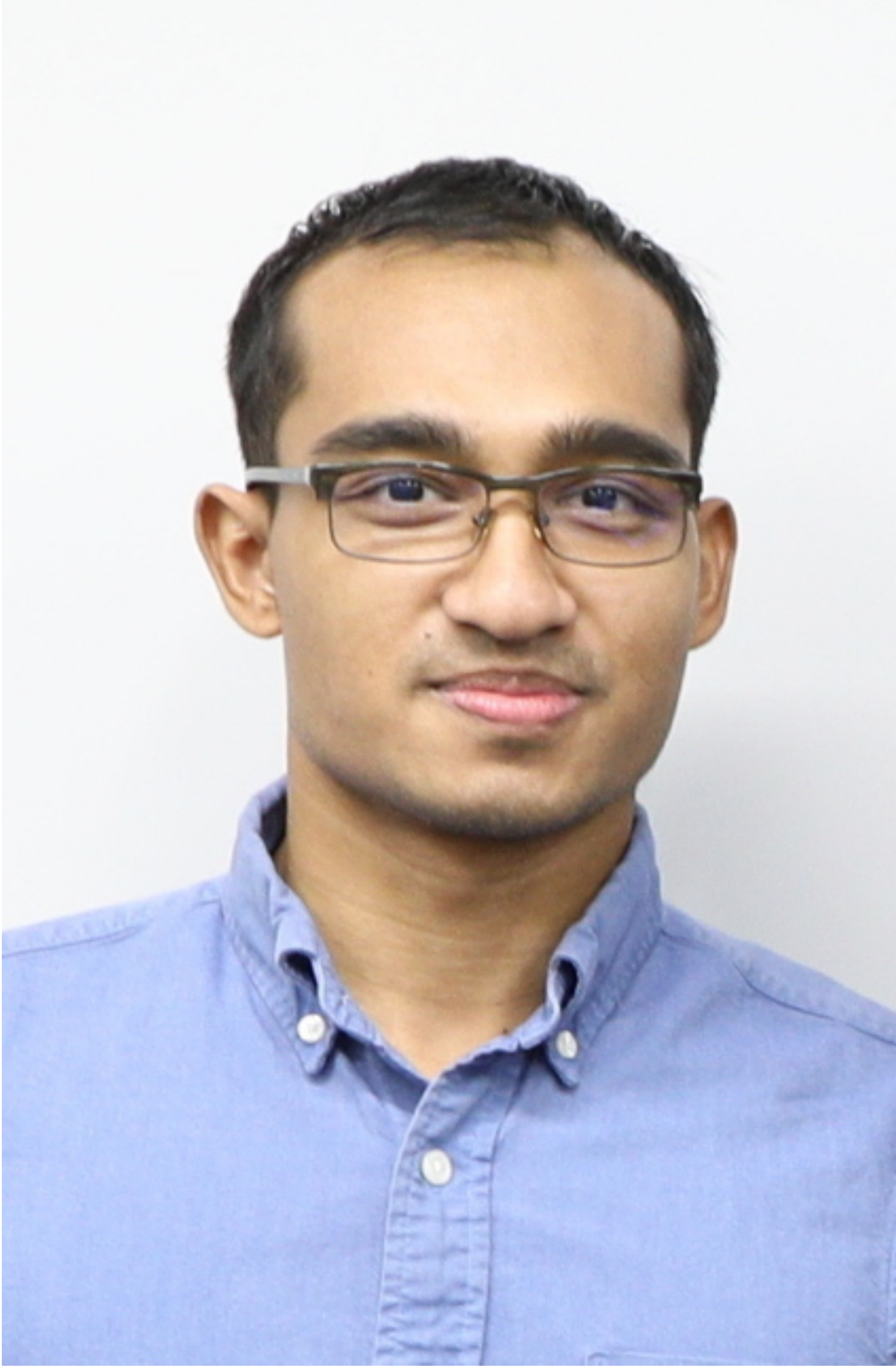]{MaungMaung AprilPyone}{received his B.C.S degree from the International Islamic University Malaysia in 2013 under the Albukhary Foundation Scholarship, M.C.S degree from the University of Malaya in 2018 under the International Graduate Research Assistantship Scheme, and Ph.D. degree from the Tokyo Metropolitan University in 2022 under the Tokyo Human Resources Fund for City Diplomacy Scholarship. He received an IEEE ICCE-TW Best Paper Award in 2016. His research interests are in the area of adversarial machine learning and information security. He is a graduate student member of IEEE.}

\profile[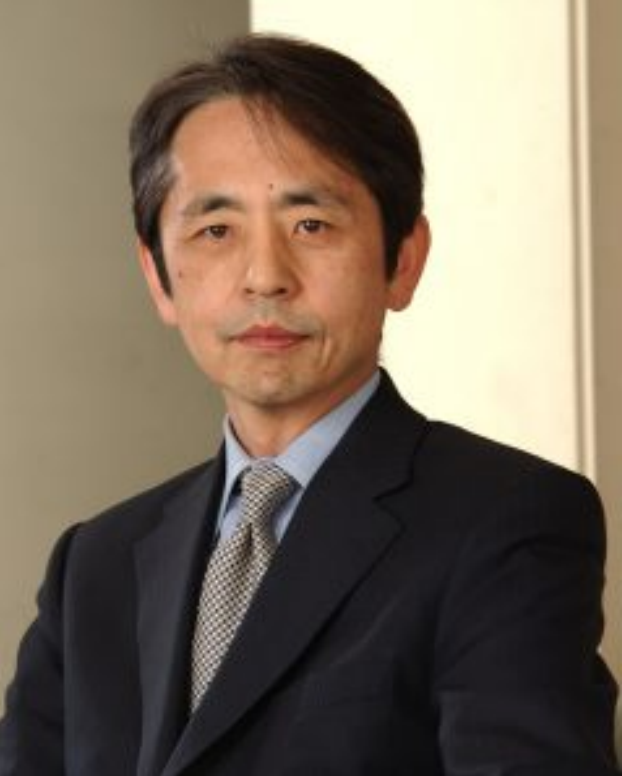]{Hitoshi Kiya}{ received B.E. and M.E. degrees from the Nagaoka University of
Technology, Japan, in 1980 and 1982, respectively, and a Dr.Eng. degree from Tokyo Metropolitan University in 1987. In 1982, he joined Tokyo Metropolitan University, where he became a Full Professor in 2000. From 1995 to 1996, he attended The University of Sydney, Australia, as a Visiting Fellow. He is a fellow of IEEE, IEICE, and ITE. He served as the President of APSIPA from 2019 to 2020 and the Regional Director-at-Large for Region 10 of the IEEE Signal Processing Society from 2016 to 2017. He was also the President of the IEICE Engineering Sciences Society from 2011 to 2012. He has been an editorial board member of eight journals, including IEEE TIP, IEEE TSP,
and IEEE TIFS. He has organized a lot of international conferences in such roles as the TPC Chair of IEEE ICASSP 2012 and as the General Co-Chair of IEEE ISCAS 2019.
}

\end{document}